\documentclass[runningheads]{llncs}

\usepackage{accv}

\usepackage{accvabbrv}

\usepackage{graphicx}
\usepackage{booktabs}
\usepackage{multirow}
\usepackage[para,online,flushleft]{threeparttable}
\usepackage{adjustbox} 

\usepackage[accsupp]{axessibility}  

\usepackage[pagebackref,breaklinks,colorlinks,citecolor=accvblue]{hyperref}

\usepackage{orcidlink}

\begin{document}

\title{Redefining Normal: A Novel Object-Level Approach for Multi-Object Novelty Detection} 

\titlerunning{A Novel Object-Level Approach for Multi-Object Novelty Detection}

\author{Mohammadreza Salehi \and Nikolaos Apostolikas,\\
Efstratios Gavves \and 
Cees G. M. Snoek \and
Yuki M. Asano }

\authorrunning{M.~Salehi et al.}

\institute{University of Amsterdam, The Netherlands \\
\email{s.salehidehnavi@uva.nl}}

\maketitle

\begin{abstract}

In the realm of novelty detection, accurately identifying outliers in data without specific class information poses a significant challenge. While current methods excel in single-object scenarios, they struggle with multi-object situations due to their focus on individual objects. Our paper suggests a novel approach: redefining `normal' at the object level in training datasets. Rather than the usual image-level view, we consider the most dominant object in a dataset as the norm, offering a perspective that is more effective for real-world scenarios. Adapting to our object-level definition of `normal', we modify knowledge distillation frameworks, where a student network learns from a pre-trained teacher network. Our first contribution, \textsc{defend}~(Dense Feature Fine-tuning on Normal Data), integrates dense feature fine-tuning into the distillation process, allowing the teacher network to focus on object-level features with a self-supervised loss. The second is masked knowledge distillation, where the student network works with partially hidden inputs, honing its ability to deduce and generalize from incomplete data. This approach not only fares well in single-object novelty detection but also considerably surpasses existing methods in multi-object contexts. The implementation is available at: \url{https://github.com/SMSD75/Redefining_Normal_ACCV24/tree/main}
%
\end{abstract}    
\section{Introduction}
\label{sec:intro}

The goal of novelty detection is to recognize samples at test time that are improbable to have originated from the training distribution~\cite{scholkopf1999support, ruff2018deep,  perera2021one, salehi2021unified}. Samples identified as deviating from the norm are labeled as anomalous, whereas the dataset used for training is considered normal. During the training phase, access is limited exclusively to normal data without utilizing any class label information. 
Progress on common novelty detection benchmarks, such as CIFAR-10~\cite{krizhevsky2009learning}, and CIFAR-100~\cite{krizhevsky2009learning}, has been impressive owing to employing the pre-trained features of foundational models~\cite{cohen2022transformaly}, generating fake samples through diffusion models~\cite{mirzaei2022fake}, or utilizing self-supervised techniques~\cite{reiss2023mean}. However, all these methods implicitly rely on the assumption that the normal distribution is object-centric and solely contains normal information in each input, which is due to evaluating on conventional datasets such as CIFAR-10. Hence, they will fail in more realistic multi-object settings.

\begin{figure}
    \centering
    \includegraphics[width=\columnwidth, trim=1cm 4.5cm 1cm 0cm, clip]{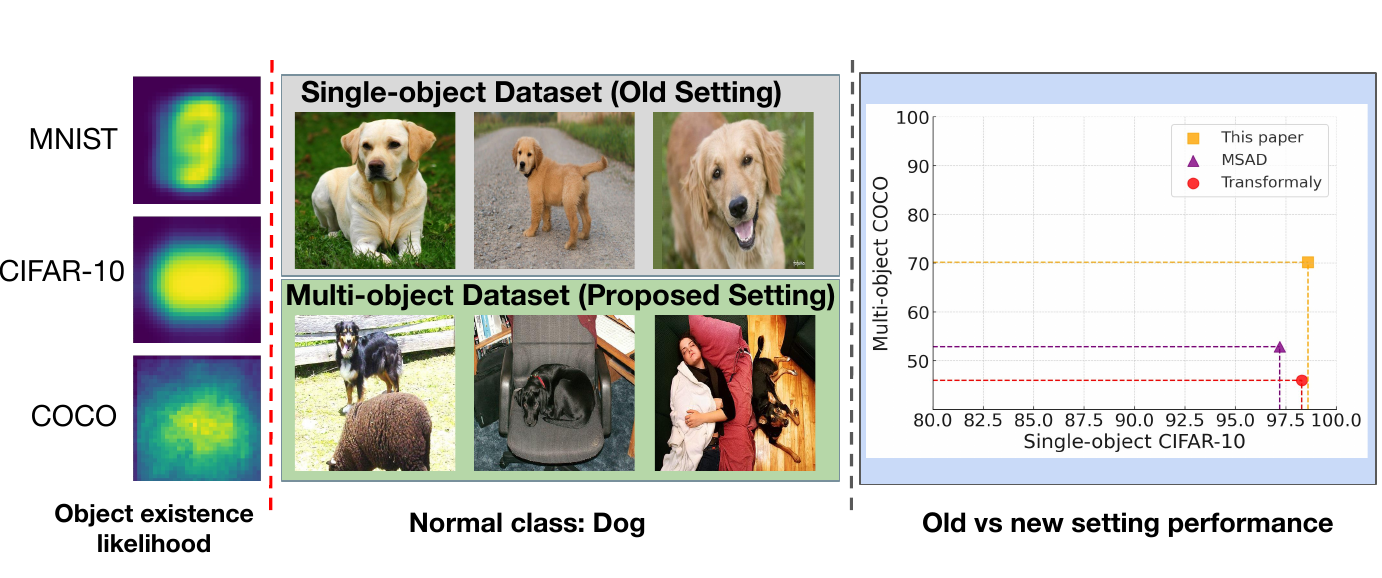}
    \caption{\textbf{A new setting and new method.} On the left, the likelihood of object existence for different datasets is shown. COCO, as opposed to MNIST and CIFAR-10, shows less object-centric biases. In the middle, we introduce a \textit{multi-object} novelty detection setting,  where we define the `normal' class as the predominant object in the dataset. In contrast to previous object-centric datasets, images can include objects of other categories (\eg, a human or a sheep for the dog class).
    On the right, we introduce a novel method that not only obtains the state-of-the-art in this setting but also excels in the classic single-object and object-centric settings. 
    }
    \label{fig:fig1}
\end{figure}

In examining state-of-the-art novelty detection methods, it appears the object-centric assumption is (unintentionally) exploited. For instance, in methods such as Transformaly~\cite{cohen2022transformaly}, KDAD~\cite{Salehi_2021_CVPR}, or RD4AD~\cite{deng2022anomaly}, a pre-trained network serves as a teacher, and a student network is trained to mimic the teacher’s output on normal training samples. The discrepancy between the teacher and student outputs is then used to detect anomalies at test time. However, the pre-trained networks are usually trained on object-centric datasets as well, hence would produce non-informative features for multi-object inputs~\cite{ziegler2022self}. This limitation also applies to self-supervised methods like MSAD~\cite{reiss2023mean} or CSI~\cite{tack2020csi}, which may focus on a single object and overlook others. Figure~\ref{fig:fig1} illustrates this issue where we apply a salient object segmentation method~\cite{simeoni2023unsupervised} on each dataset and provide the averaged foreground segmentation map locations.  In contrast, in this paper, we wish to investigate the more realistic setting, which is multi-object novelty detection. As in the real-world application of novelty detectors, it's impractical to filter test inputs to only contain normal or abnormal content; unlike controlled datasets like CIFAR-10, we propose redefining the normal training distribution to reflect the most prevalent object in a dataset. For instance, if `cat' images are dominant, this should be considered the norm, even if other objects are present in some inputs.


In our efforts to progress from single-object to multi-object novelty detection, we build upon knowledge distillation-based frameworks; yet, unlike previous methods \cite{Salehi_2021_CVPR, deng2022anomaly, cohen2022transformaly, strater2024generalad}, our approach involves a teacher model that is fine-tuned with a self-supervised loss to learn object-level features. Importantly, our self-supervised learning loss avoids the object-centric assumptions common in past methodologies~\cite{reiss2023mean, tack2020csi}. As the first major contribution,  we introduce DEnse FEature fine-tuning on Normal Data (\textsc{defend}), a self-supervised fine-tuning stage. \textsc{defend}, drawing inspiration from~\cite{ziegler2022self, salehi2023time}, shifts the focus from merely optimizing the image representation of two input views to be similar. Instead, it emphasizes ensuring that representations of any two patches are similar if they represent the same object, thus refining the fine-tuning process with a more targeted approach.
This approach offers several advantages. Firstly, it bridges the domain gap between pre-trained features and the given normal distribution, making features more distinct for different objects – a crucial aspect in scenarios involving multi-object inputs. Secondly, as noted in~\cite{ziegler2022self}, classification-based pre-trained dense features often show inconsistencies across different inputs for the same objects; for instance, patches representing dogs in two distinct images may not have identical dense feature representations. This inconsistency poses a challenge in the knowledge distillation phase for the student model, which has to predict varying targets for the same content. Aligning dense features, therefore, not only resolves this inconsistency, but also enhances the student's feature representation.

In addition, as our second contribution, we improve the robustness of the features learned by the student in the knowledge distillation by feeding it a masked input, in contrast to the teacher network that processes the complete, unaltered one. This approach boosts computational efficiency as the student network handles fewer tokens. More importantly, it forces the network to infer the teacher's knowledge from partial information, considerably improving the quality of the feature representation. Finally, we demonstrate that with appropriate feature normalization and distillation functions, our framework achieves high performance by solely distilling the last layer features, simplifying the knowledge distillation-based approaches.

Overall, this paper makes the following contributions:
\begin{itemize}
    \item We introduce DEnse FEature fine-tuning on Normal Data (\textsc{defend}), a novel self-supervised fine-tuning stage that aligns dense feature representations more effectively across different inputs, particularly benefiting multi-object scenarios.
    \item We develop an innovative knowledge distillation technique where the student network receives a masked input, enhancing computational efficiency and the quality of feature representation, leading to improved performance in the semantic novelty detection task.
    \item We propose redefining `normal' in novelty detection training datasets by focusing on the most prevalent content, introducing the first multinomial novelty detection benchmark, and evaluating state of the art methods on it.
\end{itemize}

\section{Related Works}
\label{sec:related}

\textbf{Pixel-level Anomaly Detection.} These methods usually address the problem of industrial defect detection, in which intact input samples are given during training, and the goal is to detect test time defections. One popular benchmark for such methods is MVTecAD~\cite{bergmann2019mvtec}. Two dominant approaches are embedding-based and synthesizing-based methods.
Embedding-based methods~\cite{guo2023template, liu2023simplenet, bae2023pni, gudovskiy2022cflow} focus on embedding normal image features into a compressed space to distinctly identify anomalous features, utilizing ImageNet pre-trained networks like PaDiM~\cite{defard2021padim} and PatchCore~\cite{roth2022towards}. These methods excel in distinguishing anomalous inputs at the pixel-level; yet suffer from poor performance on the semantic-level inputs.

Synthesizing-based~\cite{zhang2023destseg, li2021cutpaste, zavrtanik2021draem} methods train on synthetic anomalies to establish a boundary between normal and anomalous features. However, their reliance on unrealistic synthesized images can lead to significant deviations from normal features and a loosely defined normal feature space, making it challenging to accurately distinguish subtle anomalies. Also, such methods require specific synthetic data generators, which do not work on semantic-level datasets. 
Different from these methods, we are addressing the problem of semantic novelty detection, in which an anomaly is defined as a different semantic category.\\ 

\noindent \textbf{Semantic Novelty Detection.} These methods aim to detect novel semantic inputs that are unlikely to come from the training distribution. Self-supervised methods and pre-trained feature utilization are the leading techniques used for this problem. Self-supervised methods such as~\cite{reiss2023mean, tack2020csi, salehi2020puzzle, hendrycks2019using, golan2018deep} learn a normal representation that is further used to detect abnormal inputs. The other category of approaches leverages existing pre-trained features through adaptation~\cite{reiss2021panda, mirzaei2022fake}. They can involve two main strategies: fitting either parametric or non-parametric distributions to normal training inputs~\cite{bergman2020deep}, or training a student network. The goal of the student network is to predict the teacher's output of different layers for normal data, while failing when faced with abnormal inputs~\cite{bergmann2020uninformed, Salehi_2021_CVPR, cohen2022transformaly, deng2022anomaly}. Transformaly~\cite{cohen2022transformaly}, the most closely related lies in this category. It utilizes a knowledge distillation framework based on vision transformers and achieves state-of-the-art results. Our method, however, diverges in several significant ways. Firstly, unlike Transformaly where the teacher model remains static and unchanged throughout the training, we introduce a refinement step where the teacher is fine-tuned to better handle multi-object inputs. Secondly, we propose a novel adaptaion of the usual knowledge distillation loss by introducing `guided masked knowledge distillation'. This advancement not only markedly enhances performance but also elevates the computational efficiency of the model. Finally, we simplify the architecture and show distilling only the last layer is enough if appropriate feature normalization is applied. 

All the explained methods have primarily been assessed using datasets focused on single objects, often presupposing a singular dominant category or object in the input. As we will show, this assumption results in diminished performance when applied to multi-object datasets featuring multiple objects. To overcome this limitation, our approach introduces a novel framework that pushes the current state-of-the-art methods to effectively accommodate multi-object datasets. 
\section{Method}
\label{sec:method}

Our method primarily addresses two key challenges: firstly, learning object-level features for a normal distribution that potentially includes multi-object images, and secondly, utilizing these features to detect abnormal inputs effectively. To tackle the first challenge, we introduce \textsc{defend}, a DEnse FEature fine-tuning approach using Normal Data. This self-supervised module focuses on learning patch-level features rich in object-level information, which are represented through the patches.

We begin with a pre-trained network and fine-tune its last two layers. This process is designed to create an output feature space where object-level features are encoded at corresponding spatial locations. This approach contrasts with the initial state, where all information was encoded in a single embedding, namely the `classification token' of vision transformers.

\begin{figure*}
    \centering
    \includegraphics[width=1\textwidth, trim=0cm 10cm 0cm 0cm, clip]{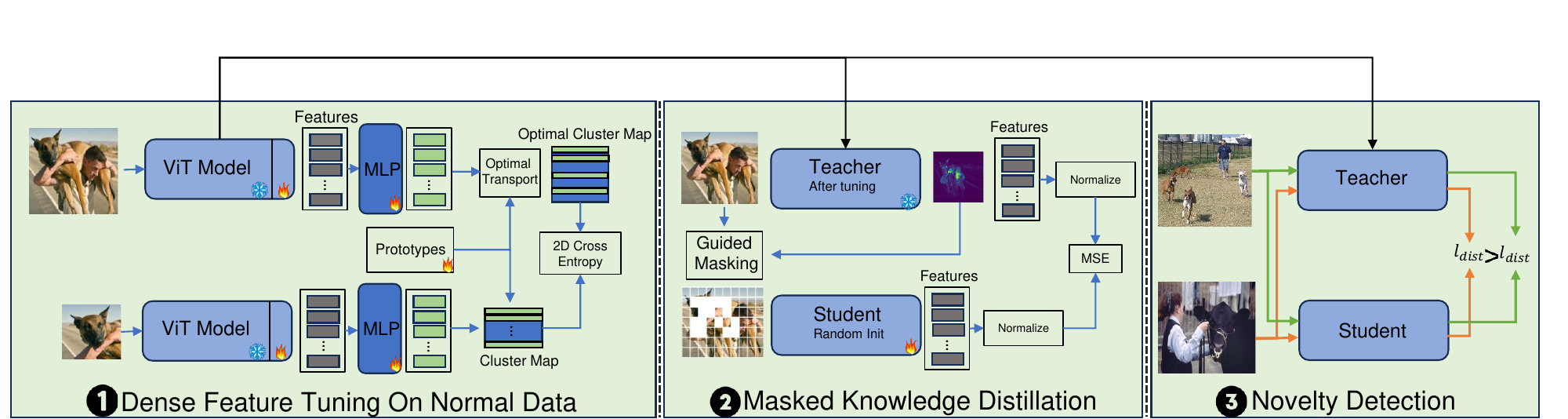}
    \caption{\textbf{The proposed method overview.} In the first stage, the last two layers of the pre-trained feature are fine-tuned on the inputs using a dense self-supervised loss to provide consistent spatial features, output features except the CLS token, for different object views. This is done by projecting the spatial features by a shared MLP head and making the corresponding features similar while avoiding trivial solutions from happening. In the second stage, the knowledge distillation framework is employed, except the input is masked by the guidance of the teacher's attention map. The student's input is masked over the most informative regions by 50\%.  Finally, the discrepancy between the student and teacher is used for the novelty detection task at the test time.}
    \label{fig:main_fig}
\end{figure*}

Next, to address the second challenge, we adopt a knowledge distillation-based framework. Here, we distill the information of normal objects into a randomly initialized student model. The student is designed to emulate the teacher's output on normal test-time inputs and to deliberately fail to do so for abnormal inputs. To prevent the student model from overfitting to shortcut solutions, we introduce a novel approach: masking the student's input, guided by the teacher, as opposed to using the unaltered input provided to the teacher. This method further encourages the student to concentrate on object-level features. The overall architecture of our proposed method is depicted in Figure~\ref{fig:main_fig} and is elaborated upon in the subsequent sections.



\subsection{Dense Feature Tuning On Normal Data}

The primary goal of this fine-tuning is to enable the network to generate consistent and meaningful semantic patch-level feature representations. These representations should be similar for various inputs only if they portray identical objects or object parts. By doing so, we significantly improve the network's proficiency in object-level novelty detection tasks.

Consider an image represented as $x \in \mathbb{R}^{3 \times H \times W}$. From this image, a random crop, denoted as $x' \in \mathbb{R}^{3 \times H' \times W'}$, is extracted. Subsequently, both $x$ and $x'$ are decomposed into patches of size $P \times P$, resulting in $N {=} \frac{H}{P} {\times} \frac{W}{P}$ patches for $x$ and $M {=} \frac{H'}{P} {\times} \frac{W'}{P}$ patches for $x'$, labeled as $x_i, i \in {1, \ldots, N}$, and $x'_j, j \in {1, \ldots, M}$, respectively. These patches are then processed through a pre-trained encoder $\Psi$, generating a set of spatial tokens and one classification token. Since the spatial features contain more object-level features compared to the classification token, we solely focus on fine-tuning the spatial features, expressed as $\Psi(x) {=} [{\Psi(x_1), \ldots, \Psi(x_N)}]$ for $x$ and $\Psi(x') {=} [{\Psi(x'_1), \ldots, \Psi(x'_M)}]$ for $x'$. The goal is to make $\Psi(x')$ similar to the features in $\Psi(x)$ for the overlapping regions in the two images. 
However, inducing the feature similarity naively can lead to trivial solutions where all spatial features become indistinctly similar. To prevent this, we first initialize a set of $K$ learnable random prototypes $\mathbf{p} \in \mathbb{R}^{K \times D}$, similar to~\cite{ziegler2022self}. Following this, the spatial tokens are processed through an MLP projection head $g$, which is equipped with L2-normalization. This procedure generates image projection features $z \in \mathbb{R}^{N \times D}$ and crop projection features $z' \in \mathbb{R}^{M \times D}$.

Crucially, instead of directly aligning individual pairs of features $(z_i, z'_j)$, our focus shifts to align the distributions of their similarities with the prototypes. This process can be seen as a matching problem, which can be addressed by optimal-transport~\cite{cuturi2013sinkhorn}, utilized in works such as SeLa~\cite{asano2020self} and SwAV~\cite{caron2020unsupervised}. This ensures a more nuanced and effective matching, which facilitates a balanced and equitable assignment of features to prototypes, avoiding lopsided or trivial associations that could impair the model's discriminative power and resulting in an optimal cluster-map $Q \in \mathbb{R}^{N \times K}$. However, as gradients cannot be directly propagated through the cluster-map $Q$, we determine the similarity distribution between the crop projection features and the prototypes, denoted as $Q' \in \mathbb{R}^{M \times K}$, through the computation of cosine similarity. This step ensures that the model maintains its learning capability while achieving a comprehensive representation of feature relationships.

Having found the optimal cluster-map $Q'$ and the cluster-map $Q$, the goal is to make corresponding regions similar. To do so, we define $\alpha(\cdot)$ as a function that gets $Q$ and returns the $M$ patches that correspond to the area from which the crop is taken. Now, the final loss function is defined as follows: 
\begin{equation}
l_{\text{dense}}(x, x') = H(Q', \alpha(Q)) = H\left(g(\Psi(x'))^T \mathbf{p}, \alpha(Q)\right),
\end{equation}
\noindent where $H$ is a 2D cross-entropy loss function.

\subsection{Masked Knowledge Distillation}
Having crafted a dense feature space that contains object-level information of normal inputs, we need another module that can detect whether a normal object is present in a given input. To do so, we employ knowledge distillation methods to distill the knowledge of normal objects into a randomly initialized student network. Intuitively, a student trained solely on normal distributions is able to emulate the teacher's output for normal test-time inputs; yet, it fails to do so for abnormal ones. To enhance the student's feature space, we mask its input, guided by leveraging the teacher's attention map to concentrate more on the informative regions:
\begin{align}
    F_{t}, \textsc{Attn}_{t} & = \Psi_{t}(x)  \notag \\
    x_{\text{masked}} &  = \text{Mask}(x, \textsc{Attn}_{t}),  
\end{align}
where  \textsc{Attn} is the Vision Transformer's attention map with regard to its $\textsc{[CLS]}$ token. 
Next, the input and the masked one are passed to the teacher and student to produce the last layer's embeddings, $F_{s}  = \Psi_{s}(x_{\text{masked}})$, which are subsequently both $L_2$ normalized: 

\begin{align}
    F_{s}^{\text{Norm}} &= \text{L2-Norm}(F_{s}) \notag \\ 
    F_{t}^{\text{Norm}} & = \text{L2-Norm}(F_{t}). \notag  \\
\end{align}

Finally, the loss is given by the Squared Error between the student's and the teacher's normalised features:

\begin{align}
    l_{\text{dist}} =  \left\| F_{t}^{\text{Norm}} - F_{s}^{\text{Norm}}\right\|^2. 
\end{align}

This loss can be interpreted as forcing the student to infer the full context by only partially observing the input, which is beneficial from several angles. 
Firstly, by requiring the student to infer the teacher's output with reduced context, it avoids the student's reliance on shortcuts, thus enhancing the representation of normal features.
Secondly, considering the distillation phase involving two forward passes, using large teacher and student architectures naively could become problematic due to the self-attention operations of ViTs scaling quadratically in memory with the number of input patches.
Masking emerges as an effective solution to this challenge, reducing memory and increasing speed. Note that, as opposed to ~\cite{cohen2022transformaly, Salehi_2021_CVPR}, which need to distill multiple layers, our model only needs to distill the last layer, further increasing simplicity and training speed.

\subsection{Computing Novelty Score}
At test time, every input $x$ is processed by both the teacher and the student networks to calculate $l_{\text{dist}}$, which is then considered as the novelty score. Novel inputs are subsequently identified based on a predefined threshold tailored to the specific use case. For evaluation, this threshold is varied and AUROC calculated.

\section{Experiments}
\label{sec:experiment}

We describe implementation details, the benchmark settings and datasets that we used, and present the results of
the experiments in the following sub-sections.

\subsection{Datasets}
Here, we briefly explain the datasets used in our experiments.\\
\noindent \textbf{CIFAR-10 and CIFAR-100}~\cite{krizhevsky2009learning}: These datasets are benchmarks in the field of novelty detection. CIFAR-10 comprises 60,000 32x32 color images in 10 classes, while CIFAR-100 has the same number of images spread over 100 classes.\\
\textbf{MNIST}~\cite{deng2012mnist}: A classic dataset containing 70,000 28x28 grayscale images of handwritten digits (0-9), widely used in the field so far.\\
\textbf{Fashion-MNIST}~\cite{xiao2017fashion}: Providing a more challenging benchmark than MNIST, this dataset includes 70,000 28x28 grayscale images across 10 fashion categories.\\
\textbf{Pascal VOC12}~\cite{everingham2015pascal}: A prominent dataset in object detection, containing a rich set of images annotated for classification, detection, segmentation.\\
\textbf{COCO}~\cite{lin2014microsoft}: Known for its complexity and variety, the Microsoft COCO dataset is a large-scale benchmark for object detection, segmentation, and captioning.\\
\textbf{MVTecAD}~\cite{bergmann2019mvtec}: Specialized in industrial applications for anomaly detection, this dataset consists of high-resolution images across various objects and textures, aimed at evaluating defect detection algorithms.

We evaluated our method on two different types of datasets, \textbf{single}-object and \textbf{multi}-object. For single-object datasets, we use CIFAR-10~\cite{krizhevsky2009learning}, CIFAR-100~\cite{krizhevsky2009learning}, MNIST~\cite{deng2012mnist}, and Fashion-MNIST~\cite{xiao2017fashion}, which are commonly used. For the multi-object, we use Pascal VOC12~\cite{everingham2015pascal} and COCO~\cite{lin2014microsoft}, which are widely used in the object-detection domain as multi-object datasets yet have not been explored in this domain so far. Our method works on training datasets where normal objects dominate, which is achieved in single-object datasets by selecting all the images of a specific class, and in multi-object datasets by selecting all the images containing a specific object. This helps DEFEND learn richer representations of normal objects, leading to improved anomaly detection performance. We argue that evaluating state-of-the-art methods on multi-object datasets is crucial for making more real-world novelty detectors, since most real-world scenarios usually have multiple objects involved~\cite{hendrycks2019scaling}.  Also, for the sake of completeness, we evaluate our method on MVTecAD~\cite{bergmann2019mvtec} as a famous industrial defect detection dataset as well.

\subsection{Training Details}
We use AdamW optimizer with a learning rate equal to 1e-5 for the last two layers of the backbone and 1e-4 for the MLP head and prototypes in the dense feature fine-tuning phase. The MLP
head consists of three linear layers with hidden dimensionality of 2048 and Gaussian error linear units as activation function~\cite{hendrycks2016gaussian} and the output dimensionality is 256. We use temperature scaling in our cross-entropy loss functions with temperature equal to 0.1. We do not use any augmentations during the dense tuning, except cropping. For the knowledge distillation phase, no augmentation is used. The crops used during dense feature fine-tuning are resized to 96.

\subsection{Implementation Details}

We utilize the Vision Transformer (ViT)~\cite{dosovitskiy2020image} as the backbone. We have discovered that pretraining variations and input size choices can lead to performance differences of up to 10\%, a fact that has been recently explored in~\cite{heckler2023exploring}. Hence, for uniformity across our experiments, we adopt supervised pretraining on ImageNet-21k with an input size of 224. Also, to have a fair comparison, we apply the same modification to other methods such as Transformaly~\cite{cohen2022transformaly}, which uses a different pretraining and input size 384. For ablation studies, we use ViT-S16, and for state-of-the-art comparisons, ViT-B16 has been used similarly to \cite{mirzaei2022fake, cohen2022transformaly}. We observed that for ViT-B16 with supervised pre-training, dense-tuning is mainly effective in multi-class and multi-object datasets since the masked knowledge distillation already gets very high single-object results. However, it is still effective for other pre-trainings. The dense tuning phase is done for 3 epochs for all the datasets except COCO with 10 epochs and the masking ratio is set to 50\% for image-level experiments as we found it an optimal point. We train the knowledge distillation framework for 10 epochs. At last,  we set the number of prototypes to be twice the number of observed objects in each experiment, as suggested by SwAV~\cite{caron2020unsupervised}. For ablation studies, the number of prototypes is set to 5.

\subsection{Evaluation Settings}

We examine our method in the uni-class and multi-class settings for both single-object and multi-object datasets. 
For the uni-class scenario, we randomly designate one semantic class as normal and categorize the remainder as abnormal. During training, the model is only exposed to the normal class’ samples. In the multi-class case, a partition of the classes is considered as normal and the rest as abnormal. During the test time, the goal is to distinguish samples that are similar to the training distribution as normal. Note that for the multi-object dataset, it is ensured that all the training samples have the normal class; however, they may also contain other objects sometimes revealed in the background as well. This is closer to the real-world setting since, in the data-gathering process, it is usually hard to filter the training set to only include specific content, such as CIFAR-10.

\begin{table*}[!tb]
    \caption{\textbf{Ablation studies of architectural changes.} All the numbers are AUROC. The performance for one semantic dataset, CIFAR-10  (C10) and one pixel-level dataset, MVTecAD (MV) is reported. In Table~\ref{table:backbone} we show the effectiveness of replacing changing the backbone $\Phi$ of VGG-16 with vision transformers. In Table~\ref{table:num_layers} we show that Distilling the final layers is better for semantic datasets while not optimal for pixel-level ones. In Table~\ref{table:normalization} we show that applying $L_2$ normalization (norm.) on the features gets the best results for CIFAR-10 (C10). In Table~\ref{table:mask} we show that applying the mask to the modified architecture is beneficial for both semantic-level and pixel-level datasets and becomes even more effective when guided by the teacher. } 

    \label{Table:all_ablations}
\setlength{\tabcolsep}{2pt}
    \centering
    \begin{subtable}[t]{.23\textwidth}
        \centering
\caption{Number of layers}
\label{table:num_layers}
    {\begin{tabular}{crr}
    \toprule
    Layer & C10 & MV \\
    \toprule
    Top5 & 89.6 & 86.8\\
    Top3 & 90.8 & 86.1  \\
    Top1 & 91.4 & 85.3  \\
    \bottomrule
    \end{tabular}}
    \end{subtable}
    \hfill
    \begin{subtable}[t]{0.23\textwidth}
        \centering
\caption{Normalization}
\label{table:normalization}
    {\begin{tabular}{crr}
    \toprule
    Norm. & C10 & MV \\
    \toprule
    - & 91.4 & 85.3  \\
    Layer & 91.5 & 87.1  \\
    $L_2$& 92.6 & 87.4  \\
    \bottomrule
    \end{tabular}}
    \end{subtable}
    \hfill
        \begin{subtable}[t]{0.23\textwidth}
        \centering
        \centering
\caption{Backbone $\Phi$}
\label{table:backbone}
    {\begin{tabular}{crr}
    \toprule
    $\Phi$ & C10 & MV \\
    \toprule
    VGG-S16 & 87.2 & 87.7  \\
    ViT-16 & 89.6 & 86.8\\
    \bottomrule
    \end{tabular}}
    \end{subtable}
    \hfill
        \begin{subtable}[t]{0.22\textwidth}
        \centering
        \centering
\caption{Masking}
\label{table:mask}
    {\begin{tabular}{crr}
    \toprule
    Mask & C10 & MV \\
    \toprule
    Without & 92.6 & 87.4\\
    Random & 93.0 & 87.7\\
    Guided & 93.4 & 88.1\\
    \bottomrule
    \end{tabular}}
    \end{subtable}
\end{table*}

\subsection{Models and Baselines}
We compare our method against FYTMI~\cite{mirzaei2022fake}, which is state-of-the-art, particularly in near-distribution novelty detection. The main advantage of this method is employing diffusion models~\cite{yang2022diffusion} to generate fake abnormal inputs, which are then utilized to boost the performance. Since the generation process is expensive, we could only re-run the method on small datasets such as MNIST. Also, we compare against state-of-the-art methods that use knowledge distillation such as Transformaly~\cite{cohen2022transformaly}, KDAD~\cite{Salehi_2021_CVPR}, and RDAD~\cite{deng2022anomaly} as well as self-supervised learning methods such as MSAD~\cite{reiss2023mean}. We noticed that while MSAD is self-supervised, it works best when initialized with pre-training. We compare with feature adaptation methods such as PANDA~\cite{reiss2021panda}, and the methods that are particularly designed for multi-class novelty detection such as UniAD~\cite{you2022unified}. Also, to show the effectiveness of our method, we compare it with semi-supervised methods such as FCDD~\cite{liznerski2020explainable} and FCC-OE~\cite{liznerski2020explainable}, which employs outlier exposure in the multi-class setting as well.

\subsection{Ablation Studies}
Here, we show the effectiveness of each component in our method. We start from the original KDAD~\cite{Salehi_2021_CVPR} framework and step-by-step change it to our transformer-based architecture augmented with guided attention masking self-supervised loss. We validate the performance boost on CIFAR-10 and MVTecAD datasets as representatives of image-level and pixel-level novelty detection. Finally, we will show the effectiveness of dense feature fine-tuning for two different pre-trainings on CIFAR10. The results are shown in Table~\ref{Table:all_ablations} and Table~\ref{table:dense_tuning_effect}.\\

\noindent \textbf{Architectural changes.} We begin by replacing the traditional backbone~\cite{Salehi_2021_CVPR} with a transformer architecture, employing an $L_2$ (MSE) loss function on the top 5 blocks. This modification results in a structure akin to Transformaly, enhancing performance by approximately 2.4\% on the image-level dataset. However, it shows a slight decrease, about 1\%, in the pixel-level tasks. Subsequently, we delve into the impact of various loss functions and the effect of normalization on the final layer, analyzing their performance across both datasets. Our experiments reveal that for semantic datasets, the optimal outcome is achieved by exclusively applying the mean square error to the last layer, especially when $L_2$ normalization is used for both teacher and student models. In contrast, pixel-level tasks benefit more from employing the smooth $L_1$ loss function~\cite{girshick2015fast}, with normalization applied solely to the teacher features. As our task is semantic novelty detection, we have only included ablation studies that are relevant to this task, yet for the full tables please refer to the appendix.   

Lastly, as the table shows, both guided and random masking show a marked improvement in results. Guided masking, in particular, slightly outperforms random masking. Notably, the use of masks not only enhances performance but also reduces the computational cost of training by 50\%. This efficiency is increasingly significant given the growing size of pre-trained models used in teacher systems. Efficient utilization of these large models is becoming a paramount concern in our field.\\

\noindent \textbf{Dense fine-tuning.} We report the performance of dense fine-tuning for every class of CIFAR-10 as Table \ref{table:dense_tuning_effect} presents. For completeness, we test with both DINO~\cite{caron2021emerging} and supervised pretrainings. The proposed method consistently improves the performance by 2.1\% and 2.5\% for the two pre-trainings on average. This shows the effectiveness of the dense self-supervised learning loss for novelty detection.  

\begin{table*}[!htb]
\centering
\caption{\textbf{AUROC score of the multi-class, all-vs-one, setting.} We compare our method (rightmost column) against the alternative. * shows that the result is reported by us. All the methods have supervised pre-training. This paper generally shows a better performance compared to others. While the proposed method gets slightly worse results on the single-object datasets, it passes other methods by a large margin on multi-object datasets, which can be attributed to better understanding of object-level information compared to others.}
\label{table:sota_multiclass_fmnist_cifar_mnist}
\resizebox{\columnwidth}{!}{\begin{tabular}{lccccccc}
\toprule
\textbf{Method} & KDAD & RDAD & FITYMI & MSAD & Transformaly & This paper \\
\textbf{Backbone} & VGG-16 & WideRes-50 & ViT-B16-224 & ViT-B16-224 & ViT-B16-384/224 & ViT-B16-224 \\
\bottomrule
CIFAR10 & 51.7* & 60.1* & 82.1*  & 85.3 & \textbf{90.4}/85.5* & 84.3 \\
MNIST & 71.4* & 75.6* & 74.8* & 74.2*  & 78.8* & \textbf{86.0} \\
FMNIST & 66.5* & 75.0* & 71.6* & 72.3 & 72.5/73.7* & \textbf{77.6} \\
Pascal VOC &  58.1* & 40.4* & -  &  48.0* & 54.1* & \textbf{63.0}\\
COCO &  52.9* & 51.6* & - &  52.9* & 46.0* & \textbf{70.2}\\
\bottomrule
\end{tabular}}
\end{table*}

\begin{table*}[!htb]
\centering
\caption{\textbf{Ablation studies of applying dense fine-tuning.} All the numbers are AUROC. The backbone is ViT-S16 with supervised pre-training. MKD shows the modified knowledge distillation framework. Each column shows the normal class in the one-vs-all setting for CIFAR-10. As it is shown, \textsc{defend} improves the novelty detection performance consistently across two different pre-trainings.}
\label{table:dense_tuning_effect}
\setlength{\tabcolsep}{3pt}
\begin{tabular}{cccccccccccccc}
\toprule
\textbf{Pertaining} & \textbf{Method} & 0 & 1 & 2 & 3 & 4 & 5 & 6 & 7 & 8 & 9 & AVG \\
\midrule
\multirow{2}{*}{Supervised} & MKD & 94.1 & 96.7 & 89.0 & 80.9 & 93.1 & 87.5 & 96.7 & 95.5 & 96.2 & 96.5 & 92.6 \\
& + \textsc{defend} & 95.6 & 97.5 & 93.2 & 84.2 & 95.3 & 90.8 & 97.7 & 97.6 & 97.6 & 97.7 & \textbf{94.7} \\
\toprule
\multirow{2}{*}{DINO} & MKD & 89.4 & 92.1 & 79.3 & 65.8 & 88.3 & 82.0 & 90.8 & 88.0 & 92.4 & 93.5 & 86.1 \\
& +\textsc{defend} & 92.9 & 96.0 & 83.8 & 70.7 & 90.3 & 80.3 & 92.8 & 90.7 & 93.9 & 94.5 & \textbf{88.6} \\
\bottomrule
\end{tabular}
\end{table*}

\subsection{Comparison to the state of the art}
Here, we compare the performance of this paper with the state-of-the-art in uni-class and multi-class settings. Furthermore, by evaluating state-of-the-art methods on multi-object datasets, we provide the first benchmark for multi-object novelty detection on Pascal VOC and COCO datasets. We use the official repositories provided by the methods to report the numbers.\\




\begin{table*}[!htb]
\centering
\caption{\textbf{AUROC score of the single-class, one-vs-all, setting.}  * shows  results reproduced by this paper. All methods utilize supervisedly pre-training backbones. We find that our proposed method closely matches the state of the art on object-centric datasets such as CIFAR-10 and FMNIST. However, on multi-object datasets, we set a strong new state-of-the-art that is roughly 4\% and 8\% higher on Pascal and COCO datasets, respectively.}
\label{table:sota_results_one_class}
\setlength{\tabcolsep}{4pt}
\resizebox{\columnwidth}{!}{\begin{tabular}{lcccccccccc}
\toprule
 \textbf{Method}& RDAD & CSI & FITYMI & KDAD & MSAD & Transformaly & This paper\\
\textbf{Backbone} & WideRes-50 & ResNet-18 & ViT-B16-224 & VGG-16 & ViT-B16-224 & ViT-B16-384/224 & ViT-B16-224\\
\textbf{Pre-training} & Supervised & Random & Supervised & Supervised & Supervised & Supervised & Supervised\\
\bottomrule
CIFAR10 & 86.1 & 94.3 &\textbf{99.1} & 87.2 & 97.2 & 98.3/94.9* & \underline{98.6} \\
CIFAR100 &  - & 89.6 & \textbf{98.1} & 80.6 & 96.4 & 97.3/93.0* & \underline{97.4} \\
FMNIST & 95.0 & 94.2 & 80.5* & \textbf{94.5} & 94.2 & 94.4/92.7* & \underline{94.4} \\

Pascal VOC & 58.6* & - & - & 82.8* & 91.8* & 82.5* & \textbf{95.4} \\
COCO Detection  & 47.9* & - & - & 75.4* & 86.7* & 75.4* & \textbf{94.5} \\

\bottomrule
\end{tabular}}
\end{table*}

\noindent \textbf{Single-object uni-class novelty detection.} We present the results in Table~\ref{table:sota_results_one_class}. As it is shown, this paper gets comparable results on single-object datasets; yet it doesn't rely on expensive approaches such as generating outliers using diffusion models, employed in FITYMI~\cite{mirzaei2022fake}. Also, our method is substantially simpler compared to Transformaly by only distilling last layer's knowledge compared to 10 layers used in their approach. Note that Transformaly gets even lower results by only distilling the last layer, which has also been reported in the paper~\cite{cohen2022transformaly}. \\

\begin{table*}[!htb]
    \centering
    \caption{\textbf{AUROC score of the multi-class, half-vs-half, on CIFAR-10.} Each row shows the indices that are selected as the normal split.  * indicate results reproduced by this paper. All methods utilize supervisedly pre-training backbones. This paper surpasses the state-of-the-art by a large margin of roughly 5\% compared to the second-best-performing method. This can be attributed to better covering different normal modalities due to applying dense feature fine-tuning loss.}
    \label{table:sota_results_multiclass_cifar}
    \setlength{\tabcolsep}{4pt}
    \resizebox{\columnwidth}{!}{\begin{tabular}{ccccccccc}
    \toprule
    &\textbf{Method} & FCDD  & FCDD+OE & PANDA & KDAD  & UniAD & Transformaly & This paper \\
    &\textbf{Backbone} & - & - & ResNet-152 & VGG-16 & EfficientNet-b4 & ViT-B16-224 & ViT-B16-224 \\
    &\textbf{Pre-training} & Random & Random & Supervised & Supervised & Supervised & Supervised & Supervised\\
    \midrule
    \multirow{3}{*}{\rotatebox[origin=c]{90}{Normal set }} &{01234} & 55.0 & 71.8 & 66.6 & 64.2 & 84.4& 78.1*& 94.7\\
    &{56789} & 50.3 & 73.7 & 73.2 & 69.3 & 80.9& 92.7*& 90.9\\
    &{02468} & 59.2 & 85.3 & 77.1 & 76.4 & 93.0& 92.3*& 92.2\\
    &{13579} & 58.5 & 85.0 & 72.9 & 78.7 & 90.6& 90.7*& 95.9\\
    \midrule
    & Mean  & 55.8 & 78.9 & 72.4 & 72.1 & 87.2 & 88.4*& \textbf{93.4}\\
    \bottomrule
    \end{tabular}
    }
\end{table*}

\noindent \textbf{Single-object multi-class novelty detection.} We cover two different sets of experiments for this setting. For the first one, we sample 5 classes of CIFAR-10 dataset as normal distribution and the rest as abnormal. We repeat this experiment for four different partitioning, as is shown by Table~\ref{table:sota_results_multiclass_cifar}. This setup has been previously used by the state-of-the-art method UniAD~\cite{you2022unified}, from which we take the numbers. The results show our superiority not only over the unsupervised methods but also over FCDD, a semi-supervised method that uses outlier exposure(OE) to boost its performance. For the second experiment, following other works~\cite{reiss2023mean, cohen2022transformaly}, we consider one class as abnormal and the rest as normal, which is shown by Table~\ref{table:sota_multiclass_fmnist_cifar_mnist}. This paper improves or gets on-par results compared to the state-of-the-art in this benchmark as well. The table shows at least 4\% improvement on FMNIST and MNIST datasets, while slightly lower results on CIFAR-10. Interestingly, the results on CIFAR-10 are better than FMNIST and MNIST datasets for all the methods, regardless of being self-supervised, using pre-trained networks, or even generating outliers using generative models to boost the performance. This is surprising since MNIST is generally considered a relatively easy dataset. This observation shows that current state-of-the-art models have been optimized for datasets such as CIFAR-10 while lacking an equal generalization on other types of samples,  the potential future research direction toward creating better methods.\\

\noindent \textbf{Multi-object uni-class novelty detection.} In this experiment, we replicated the uni-class test previously detailed in Table~\ref{table:sota_multiclass_fmnist_cifar_mnist}, this time using the Pascal VOC and COCO datasets. The results demonstrate that our proposed method outperforms others in accurately identifying normal objects, showing a 4\% improvement in performance for Pascal VOC and roughly 8\% better results for COCO datasets.

We observed that when faced with multi-object inputs, MSAD tends to lose focus on the normal objects by attending to other objects to reduce the loss. Additionally, Transformaly, which relies on pre-trained teacher models exclusively trained on object-centric datasets, faces challenges in correctly detecting normal objects. \\

\noindent \textbf{Multi-object multi-class novelty detection.} In this experiment, we evaluate the extreme case in which multiple classes of a multi-object dataset are considered as the normal distribution. To make the training dataset, we select all the images that have either of a specified set of normal objects as a normal training sample, while the rest that do not have any normal objects are considered abnormal. Note that normal training samples \textit{may} contain some abnormal objects, for instance, in the background, but this is much less dominant compared to the normal objects. This setting is very close to real-world settings, which we expect to be noisier and more challenging for existing novelty detectors. 
Table~\ref{table:sota_results_multiclass_cifar} shows the results on Pascal VOC and COCO datasets. This paper surpasses the state-of-the-art by a large margin of 9\% on Pascal VOC and by 18\% on COCO. 
We demonstrate that especially for the settings that are closer to real-world scenarios, our method consistently passes strong existing methods. 
We hypothesize that the reason why MSAD fails considerably in this setting can be attributed to the self-supervised objective function that is optimized by this method. In the presence of multiple objects, the network is not required to focus on the normal object to reduce the loss. Therefore, the normal object is not properly captured by such methods. 
As for Transformaly, we observe worse performance likely simply because it has not been designed to provide meaningful features for multi-object inputs. Therefore, the normal representation has a very loose boundary, which results in close to random performance. Although our method shows a considerable improvement in this setting, the overall numbers are still far behind the other datasets that have been common in the field, such as CIFAR-10. This shows that this challenging setting offers room for further improvement by future works to develop novel multi-object novelty detection methods.

\section{Conclusion}
\label{sec:Conclusion}
\vspace{-2mm}
In this paper, we addressed the challenge of multi-object novelty detection, proposing a paradigm shift towards a broader definition of `normal' objects in training datasets. To this end, we introduced two contributions: \textsc{defend}, a dense feature fine-tuning method, and a complementary masked knowledge distillation strategy. In the first contribution, we show how dense feature fine-tuning can improve the performance of knowledge distillation-based frameworks by enriching the features with object-level information, instead of holistic image-level one. In the second contribution, we further boost the performance and efficiency of knowledge distillation by forcing the student to infer the teacher's output by only partially observing the input. Finally, we show the effectiveness of our approach by conducting comprehensive experiments on both single-object and multi-object datasets.

\noindent \textbf{Acknowledgment.} This work is financially supported by Qualcomm Technologies Inc., the University of Amsterdam, and the allowance Top consortia for Knowledge and Innovation
(TKIs) from the Netherlands Ministry of Economic Affairs
and Climate Policy.


\clearpage  

%
%
\bibliographystyle{splncs04}
\bibliography{main}

\begin{thebibliography}{10}
\providecommand{\url}[1]{\texttt{#1}}
\providecommand{\urlprefix}{URL }
\providecommand{\doi}[1]{https://doi.org/#1}

\bibitem{asano2020self}
Asano, Y.M., Rupprecht, C., Vedaldi, A.: Self-labelling via simultaneous clustering and representation learning. In: International Conference on Learning Representations (ICLR) (2020)

\bibitem{bae2023pni}
Bae, J., Lee, J.H., Kim, S.: Pni: industrial anomaly detection using position and neighborhood information. In: Proceedings of the IEEE/CVF International Conference on Computer Vision. pp. 6373--6383 (2023)

\bibitem{bergman2020deep}
Bergman, L., Cohen, N., Hoshen, Y.: Deep nearest neighbor anomaly detection. arXiv preprint arXiv:2002.10445  (2020)

\bibitem{bergmann2019mvtec}
Bergmann, P., Fauser, M., Sattlegger, D., Steger, C.: Mvtec ad--a comprehensive real-world dataset for unsupervised anomaly detection. In: Proceedings of the IEEE/CVF conference on computer vision and pattern recognition. pp. 9592--9600 (2019)

\bibitem{bergmann2020uninformed}
Bergmann, P., Fauser, M., Sattlegger, D., Steger, C.: Uninformed students: Student-teacher anomaly detection with discriminative latent embeddings. In: Proceedings of the IEEE/CVF Conference on Computer Vision and Pattern Recognition. pp. 4183--4192 (2020)

\bibitem{caron2020unsupervised}
Caron, M., Misra, I., Mairal, J., Goyal, P., Bojanowski, P., Joulin, A.: Unsupervised learning of visual features by contrasting cluster assignments. Advances in neural information processing systems  \textbf{33},  9912--9924 (2020)

\bibitem{caron2021emerging}
Caron, M., Touvron, H., Misra, I., J{\'e}gou, H., Mairal, J., Bojanowski, P., Joulin, A.: Emerging properties in self-supervised vision transformers. In: Proceedings of the IEEE/CVF international conference on computer vision. pp. 9650--9660 (2021)

\bibitem{cohen2022transformaly}
Cohen, M.J., Avidan, S.: Transformaly-two (feature spaces) are better than one. In: Proceedings of the IEEE/CVF Conference on Computer Vision and Pattern Recognition. pp. 4060--4069 (2022)

\bibitem{cuturi2013sinkhorn}
Cuturi, M.: Sinkhorn distances: Lightspeed computation of optimal transport. Advances in neural information processing systems  \textbf{26} (2013)

\bibitem{defard2021padim}
Defard, T., Setkov, A., Loesch, A., Audigier, R.: Padim: a patch distribution modeling framework for anomaly detection and localization. In: International Conference on Pattern Recognition. pp. 475--489. Springer (2021)

\bibitem{deng2022anomaly}
Deng, H., Li, X.: Anomaly detection via reverse distillation from one-class embedding. In: Proceedings of the IEEE/CVF Conference on Computer Vision and Pattern Recognition. pp. 9737--9746 (2022)

\bibitem{deng2012mnist}
Deng, L.: The mnist database of handwritten digit images for machine learning research [best of the web]. IEEE signal processing magazine  \textbf{29}(6),  141--142 (2012)

\bibitem{dosovitskiy2020image}
Dosovitskiy, A., Beyer, L., Kolesnikov, A., Weissenborn, D., Zhai, X., Unterthiner, T., Dehghani, M., Minderer, M., Heigold, G., Gelly, S., et~al.: An image is worth 16x16 words: Transformers for image recognition at scale. arXiv preprint arXiv:2010.11929  (2020)

\bibitem{everingham2015pascal}
Everingham, M., Eslami, S.A., Van~Gool, L., Williams, C.K., Winn, J., Zisserman, A.: The pascal visual object classes challenge: A retrospective. International journal of computer vision  \textbf{111},  98--136 (2015)

\bibitem{girshick2015fast}
Girshick, R.: Fast r-cnn. In: Proceedings of the IEEE international conference on computer vision. pp. 1440--1448 (2015)

\bibitem{golan2018deep}
Golan, I., El-Yaniv, R.: Deep anomaly detection using geometric transformations. Advances in neural information processing systems  \textbf{31} (2018)

\bibitem{gudovskiy2022cflow}
Gudovskiy, D., Ishizaka, S., Kozuka, K.: Cflow-ad: Real-time unsupervised anomaly detection with localization via conditional normalizing flows. In: Proceedings of the IEEE/CVF Winter Conference on Applications of Computer Vision. pp. 98--107 (2022)

\bibitem{guo2023template}
Guo, H., Ren, L., Fu, J., Wang, Y., Zhang, Z., Lan, C., Wang, H., Hou, X.: Template-guided hierarchical feature restoration for anomaly detection. In: Proceedings of the IEEE/CVF International Conference on Computer Vision. pp. 6447--6458 (2023)

\bibitem{heckler2023exploring}
Heckler, L., K{\"o}nig, R., Bergmann, P.: Exploring the importance of pretrained feature extractors for unsupervised anomaly detection and localization. In: Proceedings of the IEEE/CVF Conference on Computer Vision and Pattern Recognition. pp. 2916--2925 (2023)

\bibitem{hendrycks2019scaling}
Hendrycks, D., Basart, S., Mazeika, M., Zou, A., Kwon, J., Mostajabi, M., Steinhardt, J., Song, D.: Scaling out-of-distribution detection for real-world settings. arXiv preprint arXiv:1911.11132  (2019)

\bibitem{hendrycks2016gaussian}
Hendrycks, D., Gimpel, K.: Gaussian error linear units (gelus). arXiv preprint arXiv:1606.08415  (2016)

\bibitem{hendrycks2019using}
Hendrycks, D., Mazeika, M., Kadavath, S., Song, D.: Using self-supervised learning can improve model robustness and uncertainty. Advances in Neural Information Processing Systems  \textbf{32} (2019)

\bibitem{krizhevsky2009learning}
Krizhevsky, A., Hinton, G., et~al.: Learning multiple layers of features from tiny images  (2009)

\bibitem{li2021cutpaste}
Li, C.L., Sohn, K., Yoon, J., Pfister, T.: Cutpaste: Self-supervised learning for anomaly detection and localization. In: Proceedings of the IEEE/CVF Conference on Computer Vision and Pattern Recognition. pp. 9664--9674 (2021)

\bibitem{lin2014microsoft}
Lin, T.Y., Maire, M., Belongie, S., Hays, J., Perona, P., Ramanan, D., Doll{\'a}r, P., Zitnick, C.L.: Microsoft coco: Common objects in context. In: Computer Vision--ECCV 2014: 13th European Conference, Zurich, Switzerland, September 6-12, 2014, Proceedings, Part V 13. pp. 740--755. Springer (2014)

\bibitem{liu2023simplenet}
Liu, Z., Zhou, Y., Xu, Y., Wang, Z.: Simplenet: A simple network for image anomaly detection and localization. In: Proceedings of the IEEE/CVF Conference on Computer Vision and Pattern Recognition. pp. 20402--20411 (2023)

\bibitem{liznerski2020explainable}
Liznerski, P., Ruff, L., Vandermeulen, R.A., Franks, B.J., Kloft, M., M{\"u}ller, K.R.: Explainable deep one-class classification. arXiv preprint arXiv:2007.01760  (2020)

\bibitem{mirzaei2022fake}
Mirzaei, H., Salehi, M., Shahabi, S., Gavves, E., Snoek, C.G., Sabokrou, M., Rohban, M.H.: Fake it till you make it: Near-distribution novelty detection by score-based generative models. arXiv preprint arXiv:2205.14297  (2022)

\bibitem{perera2021one}
Perera, P., Oza, P., Patel, V.M.: One-class classification: A survey. arXiv preprint arXiv:2101.03064  (2021)

\bibitem{reiss2021panda}
Reiss, T., Cohen, N., Bergman, L., Hoshen, Y.: Panda: Adapting pretrained features for anomaly detection and segmentation. In: Proceedings of the IEEE/CVF Conference on Computer Vision and Pattern Recognition. pp. 2806--2814 (2021)

\bibitem{reiss2023mean}
Reiss, T., Hoshen, Y.: Mean-shifted contrastive loss for anomaly detection. In: Proceedings of the AAAI Conference on Artificial Intelligence. vol.~37, pp. 2155--2162 (2023)

\bibitem{roth2022towards}
Roth, K., Pemula, L., Zepeda, J., Sch{\"o}lkopf, B., Brox, T., Gehler, P.: Towards total recall in industrial anomaly detection. In: Proceedings of the IEEE/CVF Conference on Computer Vision and Pattern Recognition. pp. 14318--14328 (2022)

\bibitem{ruff2018deep}
Ruff, L., Vandermeulen, R., Goernitz, N., Deecke, L., Siddiqui, S.A., Binder, A., M{\"u}ller, E., Kloft, M.: Deep one-class classification. In: International conference on machine learning. pp. 4393--4402. PMLR (2018)

\bibitem{salehi2020puzzle}
Salehi, M., Eftekhar, A., Sadjadi, N., Rohban, M.H., Rabiee, H.R.: Puzzle-ae: Novelty detection in images through solving puzzles. arXiv preprint arXiv:2008.12959  (2020)

\bibitem{salehi2023time}
Salehi, M., Gavves, E., Snoek, C.G., Asano, Y.M.: Time does tell: Self-supervised time-tuning of dense image representations. In: Proceedings of the IEEE/CVF International Conference on Computer Vision. pp. 16536--16547 (2023)

\bibitem{salehi2021unified}
Salehi, M., Mirzaei, H., Hendrycks, D., Li, Y., Rohban, M.H., Sabokrou, M.: A unified survey on anomaly, novelty, open-set, and out-of-distribution detection: Solutions and future challenges. arXiv preprint arXiv:2110.14051  (2021)

\bibitem{Salehi_2021_CVPR}
Salehi, M., Sadjadi, N., Baselizadeh, S., Rohban, M.H., Rabiee, H.R.: Multiresolution knowledge distillation for anomaly detection. In: Proceedings of the IEEE/CVF Conference on Computer Vision and Pattern Recognition (CVPR). pp. 14902--14912 (June 2021)

\bibitem{scholkopf1999support}
Sch{\"o}lkopf, B., Williamson, R.C., Smola, A., Shawe-Taylor, J., Platt, J.: Support vector method for novelty detection. Advances in neural information processing systems  \textbf{12} (1999)

\bibitem{simeoni2023unsupervised}
Sim{\'e}oni, O., Sekkat, C., Puy, G., Vobeck{\`y}, A., Zablocki, {\'E}., P{\'e}rez, P.: Unsupervised object localization: Observing the background to discover objects. In: Proceedings of the IEEE/CVF Conference on Computer Vision and Pattern Recognition. pp. 3176--3186 (2023)

\bibitem{strater2024generalad}
Str{\"a}ter, L.P., Salehi, M., Gavves, E., Snoek, C.G., Asano, Y.M.: Generalad: Anomaly detection across domains by attending to distorted features. arXiv preprint arXiv:2407.12427  (2024)

\bibitem{tack2020csi}
Tack, J., Mo, S., Jeong, J., Shin, J.: Csi: Novelty detection via contrastive learning on distributionally shifted instances. Advances in neural information processing systems  \textbf{33},  11839--11852 (2020)

\bibitem{xiao2017fashion}
Xiao, H., Rasul, K., Vollgraf, R.: Fashion-mnist: a novel image dataset for benchmarking machine learning algorithms. arXiv preprint arXiv:1708.07747  (2017)

\bibitem{yang2022diffusion}
Yang, L., Zhang, Z., Song, Y., Hong, S., Xu, R., Zhao, Y., Zhang, W., Cui, B., Yang, M.H.: Diffusion models: A comprehensive survey of methods and applications. ACM Computing Surveys  (2022)

\bibitem{you2022unified}
You, Z., Cui, L., Shen, Y., Yang, K., Lu, X., Zheng, Y., Le, X.: A unified model for multi-class anomaly detection. Advances in Neural Information Processing Systems  \textbf{35},  4571--4584 (2022)

\bibitem{zavrtanik2021draem}
Zavrtanik, V., Kristan, M., Sko{\v{c}}aj, D.: Draem-a discriminatively trained reconstruction embedding for surface anomaly detection. In: Proceedings of the IEEE/CVF International Conference on Computer Vision. pp. 8330--8339 (2021)

\bibitem{zhang2023destseg}
Zhang, X., Li, S., Li, X., Huang, P., Shan, J., Chen, T.: Destseg: Segmentation guided denoising student-teacher for anomaly detection. In: Proceedings of the IEEE/CVF Conference on Computer Vision and Pattern Recognition. pp. 3914--3923 (2023)

\bibitem{ziegler2022self}
Ziegler, A., Asano, Y.M.: Self-supervised learning of object parts for semantic segmentation. In: Proceedings of the IEEE/CVF Conference on Computer Vision and Pattern Recognition. pp. 14502--14511 (2022)

\end{thebibliography}

\setcounter{page}{1}
\section{Additional Experiments}
\label{sec:additional_experiments}

\begin{table*}[!htb]
\centering
\caption{\textbf{Pascal VOC per class results in the uni-class setting.} This paper sets a new state-of-the-art for the proposed multi-object benchmark by considerably passing the second best-performing method. }
\label{tab:pvoc-per-class-one-class}
\begin{adjustbox}{max width=\textwidth}
\begin{tabular}{@{}ccccccc@{}}
\toprule
 & & KDAD & RDAD & MSAD & Transformaly & This paper \\ 
\midrule
\multirow{20}{*}{\rotatebox[origin=c]{90}{Normal class }} &
0 & 96.2 & 88.7 & 99.7 & 98.0 & 99.7\\
& 1 & 67.6 & 42.4 & 63.4 & 46.4 & 75.6\\
& 2 & 90.8 & 63.3 & 96.3 & 86.5 & 97.6\\
& 3 & 81.4 & 61.1 & 97.7 & 83.5 & 91.9\\
& 4 & 82.2 & 55.3 & 87.1 & 71.0 & 92.6\\
& 5 & 69.4 & 55.0 & 93.9 & 82.6 & 93.8\\
& 6 & 71.4 & 51.7 & 83.7 & 61.9 & 90.8\\
& 7 & 87.6 & 76.0 & 95.4 & 90.4 & 98.5\\
& 8 & 84.4 & 52.6 & 92.2 & 73.7 & 97.7\\
& 9 & 91.6 & 62.6 & 99.3 & 94.4 & 99.3\\
& 10 & 77.2 & 44.2 & 97.1 & 90.9 & 98.7\\
& 11 & 82.7 & 53.3 & 96.2 & 93.9 & 97.9\\
& 12 & 83.9 & 60.7 & 96.3 & 94.3 & 99.2\\
& 13 & 83.4 & 44.2 & 90.3 & 81.2 & 97.9\\
& 14 & 79.2 & 50.0 & 83.4 & 76.9 & 89.2\\
& 15 & 86.1 & 67.9 & 96.3 & 86.4 & 97.9\\
& 16 & 87.6 & 66.6 & 93.6 & 84.30 & 97.7 \\
& 17 & 95.7 & 63.6 & 98.9 & 96.2 & 98.3\\
& 18 & 72.8 & 40.4 & 77.3 & 58.1 & 92.7\\
& 19 & 90.1 & 71.5 & 98.8 & 96.8 & 99.0\\
\midrule
& AVG & 83.0 & 58.9 & 91.8 & 82.5 & \textbf{95.3}\\
\bottomrule
\end{tabular}
\end{adjustbox}
\end{table*}

\subsection{Pascal VOC per class results}
Here, we show the per-class results of both multi-class and single-class settings in Table~\ref{tab:pvoc-per-class-one-class} and Table~\ref{tab:anomaly-detection-per-class-multi-class}. The results are obtained similar to Table~2 and Table~4 in the main paper. As is shown, this paper gets better average results in both uni-class and multi-class settings for the multi-object dataset. While Transformaly and MSAD get slightly better or comparable results on single-object datasets such as CIFAR10, they show significantly lower performance on multi-object samples. This performance drop supports our postulation that these methods have implicit object-centric assumptions. 

\begin{table*}[!htb]
    \centering
    \caption{\textbf{Ablation studies of architectural changes.} All the numbers are AUROC. The performance for one semantic~(CIFAR-10) and one pixel-level~(MVTecAD) dataset is reported. As it is shown, different distillation and normalization functions get the best results for different datasets. As we are addressing semantic novelty detection, only top1 layer is distilled and $L_2$ normalization is applied on both networks. Guided masking is also used as it is consistently effective across different tasks.}
    \label{table:combined_ablations}
    \begin{threeparttable}
    \setlength{\tabcolsep}{0.2em}
    \aboverulesep=0mm \belowrulesep=0mm
    \subcaption{Ablating the effect of each modification applied to the baseline.}
    \label{table:abl_modification}
    \begin{tabular}{ccccccccc}
    \toprule
    & Backbone & Num layer & Loss & \multicolumn{2}{c}{Normalization} & & \multicolumn{2}{c}{Results} \\
    \cline{5-9}
    &          &           &      & Teacher & Student & & CIFAR-10 & MVTecAD \\
    \midrule
    \multirow{7}{*}{Modified} 
    & ViT-S16  & top5      & smooth $L_1$ & Layer & - & & 89.4 & \textbf{90.4} \\
    & ViT-S16  & top5      & $L_2$        & Layer & - & & 90.1 & 90.2 \\
    & ViT-S16  & top5      & $L_2$        & -     & - & & 89.6 & 86.8 \\
    & ViT-S16  & top3      & $L_2$        & -     & - & & 90.8 & 86.1 \\
    & ViT-S16  & top3      & $L_2$        & Layer & - & & 91.1 & 90.8 \\
    & ViT-S16  & top1      & $L_2$        & Layer & - & & 91.5 & 87.1 \\
    & ViT-S16  & top1      & $L_2$        & $L_2$ & $L_2$ & & \textbf{92.6} & 85.3 \\
    \midrule
    Baseline & VGG-16    & top5      & $L_2$ + cosine & - & - & & 87.2 & 87.7 \\
    \bottomrule
    \end{tabular}

    \vspace{1em} 
    \subcaption{Ablating the effect of masking on the top-performing models from Table~\ref{table:abl_modification}. }
    \label{table:abl_masked}
    \begin{tabular}{ccccccc}
    \toprule
    & Backbone & Num layer & Loss & Mask & \multicolumn{2}{c}{Results} \\
    \cline{6-7}
    &          &           &      &      & CIFAR-10 & MVTecAD       \\
    \midrule
    \multirow{3}{*}{With Masking} 
    & ViT-S16  & top5      & smooth $L_1$ & Random & 89.8 & 91.0    \\
    & ViT-S16  & top5      & smooth $L_1$ & Guided & 90.0 & \textbf{91.4} \\
    & ViT-S16  & top1      & $L_2$        & Guided & \textbf{93.4} & 88.1 \\
    \midrule
    \multirow{2}{*}{Without Masking} 
    & ViT-S16  & top5      & smooth $L_1$ & -      & 89.4 & 90.4    \\
    & ViT-S16  & top1      & $L_2$        & -      & 92.6 & 85.3    \\
    \bottomrule
    \end{tabular}
    \end{threeparttable}
\end{table*}

\subsection{COCO per class results}

Here, we show the per-class results for COCO with a similar evaluations as  Table~2 in the main paper. Table~\ref{Table:COCO_per_class_one_class} shows the results. As is shown, this paper nearly always gets better results compared to MSAD and Transformaly. This particularly shows that this paper is a better fit for real-world applications since most real scenarios are multi-object scenes. 

\begin{table*}[htb!]
\centering
\caption{\textbf{COCO per class results.} ID specifies the class ID in the dataset. The performance is only reported for valid class IDs. }
\label{Table:COCO_per_class_one_class}
\begin{tabular}{cccccccc}
\toprule
ID & Transformaly & MSAD & This paper & ID & Transformaly & MSAD & This paper \\ 
\midrule
0 & - & - & - & 46 & 67.2 & 70.8 & 94.4\\
1 & 46.1 & 58.2 & 77.5 & 47 & 57.0 & 89.3 & 88.1\\
2 & 56.5 & 71.5 & 88.3 & 48 & 79.5 & 75.0 & 96.2\\
3 & 57.9 & 68.2 & 86.7 & 49 & 73.6 & 93.9 & 93.6\\
4 & 82.6 & 90.4 & 96.5 & 50 & 72.0 & 91.4 & 94.5\\
5 & 93.4 & 97.8 & 98.6 & 51 & 64.5 & 90.9 & 91.6\\
6 & 79.7 & 86.1 & 95.6 & 52 & 81.2 & 83.3 & 94.8\\
7 & 91.9 & 97.0 & 98.4 & 53 & 78.9 & 90.3 & 92.9\\
8 & 63.9 & 73.2 & 87.2 & 54 & 81.0 & 90.6 & 97.3\\
9 & 74.0 & 83.8 & 95.9 & 55 & 80.3 & 95.3 & 95.7\\
10 & 79.9 & 88.8 & 96.3 & 56 & 92.9 & 91.3 & 98.3\\
11 & 59.6 & 86.6 & 95.4 & 57 & 82.3 & 97.3 & 95.2\\
12 & - & - & - & 58 & 81.7 & 93.3 & 97.8\\
13 & 76.7 & 88.2 & 95.0 & 59 & 88.9 & 96.6 & 98.5\\
14 & 83.1 & 93.8 & 97.5 & 60 & 76.0 & 98.1 & 96.1\\
15 & 56.8 & 66.3 & 79.6 & 61 & 67.2 & 94.6 & 95.9\\
16 & 62.5 & 70.5 & 82.0 & 62 & 49.2 & 92.4 & 81.0\\
17 & 65.1 & 87.2 & 97.1 & 63 & 59.8 & 65.8 & 95.9\\
18 & 44.0 & 57.6 & 82.1 & 64 & 53.0 & 89.3 & 84.9\\
19 & 75.8 & 92.8 & 95.1 & 65 & 73.4 & 68 & 97.4\\
20 & 85.1 & 96.6 & 98.5 & 66 & - & - & -\\
21 & 79.5 & 94.3 & 98.2 & 67 & 59.1 & 91.0 & 91.1\\
22 & 95.7 & 98.3 & 98.7 & 68 & - & - & -\\
23 & 94.9 & 98.9 & 98.7 & 69 & - & - & -\\
24 & 98.0 & 99.3 & 99.4 & 70 & 94.7 & 79.0 & 99.5\\
25 & 95.0 & 98.7 & 99.4 & 71 & - & - & -\\
26 & - & - & - & 72 & 74.0 & 98.9 & 96.2\\
27 & 51.8 & 64.4 & 78.6 & 73 & 77.3 & 90.1 & 97.3\\
28 & 62.1 & 70.6 & 91.6 & 74 & 91.4 & 91.8 & 98.9\\
29 & - & - & - & 75 & 56.3 & 97.3 & 95.3\\
30 & - & - & - & 76 & 90.4 & 91.7 & 98.8\\
31 & 44.8 & 58.1 & 78.8 & 77 & 43.9 & 96.6 & 85.6\\
32 & 54.8 & 75.4 & 91.9 & 78 & 85.7 & 63.7 & 98.5\\
33 & 48.1 & 75.1 & 91.0 & 79 & 83.8 & 95.4 & 98.4\\
34 & 60.5 & 92.7 & 94.3 & 80 & 89.4 & 95.2 & 98.7\\
35 & 97.7 & 99.3 & 99.4 & 81 & 87.8 & 97.0 & 98.0\\
36 & 93.7 & 97.8 & 98.3 & 82 & 79.6 & 96.4 & 97.1\\
37 & 90.0 & 94.6 & 92.8 & 83 & - & - & -\\
38 & 72.2 & 96.1 & 97.7 & 84 & 58.0 & 93.8 & 90.1\\
39 & 93.4 & 98.8 & 98.2 & 85 & 68.6 & 78.4 & 90.0\\
40 & 96.5 & 99.2 & 99.0 & 86 & 72.0 & 78.5 & 94.7\\
41 & 75.1 & 95.5 & 98.7 & 87 & 56.0 & 88.0 & 93.7\\
42 & 89.4 & 97.7 & 98.4 & 88 & 65.8 & 84.6 & 95.0\\
43 & 97.2 & 99.6 & 99.7 & 89 & 81.3 & 87.6 & 94.8\\
44 & 53.4 & 70.8 & 84.3 & 90 & 74.9 & 94.0 & 95.4\\
45 & - & - & - & & & \\
\bottomrule
\end{tabular}
\end{table*}

\subsection{Analyzing computational efficiency} The training and inference times for both Transformaly and our method are shown in Table~\ref{table:comput_compare} using an A6000 GPU. Our method includes two training stages, requiring 3 and 10 epochs respectively, leading to a slightly longer training time. Despite this, our method achieves approximately 50\% faster inference time, which is more critical in practical applications since anomaly detection models are usually trained once but used repeatedly.

\begin{table}[!thb]
    \centering
    \caption{\textbf{Computational efficiency on CIFAR-10.}}
    \resizebox{\columnwidth}{!}{\begin{tabular}{lccccccc}
        \toprule
        \textbf{Method} & \textbf{Batch size} & \textbf{Training epochs} & \textbf{Training time (min)} & \textbf{Inference FPS} &\textbf{Performance} \\
        \midrule
        Transformaly & 32 & 10 & 63 & 100 & 94.9 \\
        \hline
        Ours &  32    & 13~(3 + 10) & 75~(45 + 30) &  164 &  98.6  \\
        \bottomrule
    \end{tabular}
    }
    \label{table:comput_compare}
\end{table}

\section{Additional Ablations}
\label{sec:additional_ablations}
Here, we provide ablation studies on the effect of different normalization and distillation functions applied to the features of different layers. As shown in Table~\ref{table:abl_masked}, the smooth $L_1$~\cite{girshick2015fast} loss is the most effective distillation function for pixel-level tasks. Also, applying normalization is beneficial on both pixel-level and semantic tasks. When only the teacher network is normalized, the student's features are projected to the teacher's representation space by a shared linear head. Moreover, the results demonstrate that increasing the number of distillation layers is mainly beneficial for pixel-level and not semantic tasks.  In this paper, we focus more on distilling the final block with $L_2$ normalization applied to all the network's features and $L_2$ distillation loss, as our task is semantic novelty detection. 

In Table~\ref{table:abl_masked}, we choose the top-performing models from Table~\ref{table:abl_modification} and evaluate them in situations with different input masking procedures. As is shown, different kinds of masking are beneficial for both semantic and pixel-level tasks. Masking helps improve the results for CIFAR-10 by roughly 1.2\%  and MVTecAD by 1\% , which shows the generality of the proposed approach. Also, guided masking shows consistently better performance compared to random masking, which supports the effectiveness of giving more attention to informative areas instead of random.


\begin{table}[!thb]
\centering
\caption{\textbf{Pascal VOC per class results in the multi-class setting.} This paper pushes the state-of-the-art by roughly 5\% in the proposed multi-object benchmark.}
\label{tab:anomaly-detection-per-class-multi-class}
\begin{adjustbox}{max width=\textwidth}
\begin{tabular}{@{}cccccccc@{}}
\toprule
& & MSAD & Transformaly & RDAD & KDAD & DSVDD & \textbf{This paper}\\ 
\midrule
\multirow{20}{*}{\rotatebox[origin=c]{90}{Abnormal class }}
& 0 & 33.3 & 21.3 & 48.4 & 36.1 & 45.4 & 40.9 \\
& 1 & 44.1 & 91.1 & 60.1 & 62.4 & 48.3 & 68.0 \\
& 2 & 56.9 & 39.4 & 51.6 & 46.6 & 47.8 & 79.6 \\
& 3 & 41.5 & 67.9 & 53.6 & 71.9 & 46.1 & 85.4 \\
& 4 & 39.1 & 70.6 & 52.2 & 40.3 & 34.3 & 48.4 \\
& 5 & 51.7 & 73.9 & 56.1 & 83.9 & 43.6 & 70.2 \\
& 6 & 34.3 & 70.9 & 53.0 & 84.4 & 56.7 & 91.9 \\
& 7 & 48.2 & 51.0 & 38.7 & 49.6 & 41.2 & 60.7 \\
& 8 & 16.8 & 62.4 & 42.5 & 62.9 & 46.6 & 33.4 \\
& 9 & 51.9 & 23.9 & 42.5 & 53.6 & 47.30 & 61.7 \\
& 10 & 66.6 & 24.4 & 44.3 & 59.5 & 44.4 & 58.2 \\
& 11 & 50.7 & 51.0 & 43.2 & 55.0 & 51.0 & 60.1 \\
& 12 & 60.3 & 56.9 & 43.2 & 55.1 & 51.7 & 53.4 \\
& 13 & 59.0 & 55.7 & 44.2 & 63.8 & 58.5 & 71.5 \\
& 14 & 55.4 & 53.2 & 41.5 & 57.6 & 53.5 & 69.3 \\
& 15 & 43.8 & 64.2 & 39.9 & 72.4 & 42.1 & 83.0 \\
& 16 & 45.9 & 61.5 & 39.9 & 35.9 & 40.2 & 54.8 \\
& 17 & 58.1 & 25.8 & 39.8 & 38.2 & 44.8 & 49.0 \\
& 18 & 51.5 & 65.7 & 42.6 & 76.0 & 59.4 & 82.0 \\
& 19 & 51.4 & 51.4 & 44.6 & 56.8 & 41.9 & 56.2 \\
\midrule
& AVG & 48.0 & 54.1 & 46.2 & 58.1 & 47.3 & \textbf{63.0} \\
\bottomrule
\end{tabular}
\end{adjustbox}
\end{table}

\section{Qualitative Results}

\begin{figure}[!tbh]
    \centering
    \includegraphics[width=\columnwidth, trim=1cm 5cm 2cm 0cm, clip]{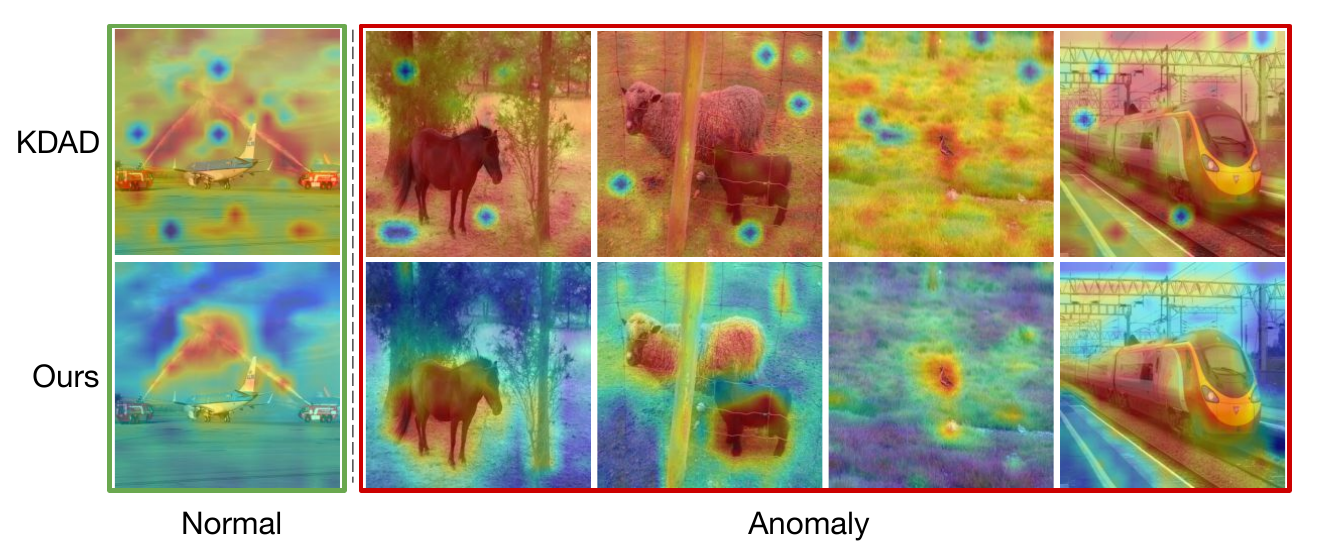}
    \caption{\textbf{Qualitative comparison of the proposed method and KDAD.} As shown, the proposed method focuses more on the abnormal object to produce the anomaly score for each image.}
    \label{fig:rebuttal_fig_1}
\end{figure}

We are comparing our method with KDAD~\cite{Salehi_2021_CVPR} in terms of the regions each method focuses on to calculate the anomaly score for each input image. In this comparison, the normal class is `Airplane', and we expect the anomaly score to be lower for the areas containing airplanes and higher for regions containing abnormal objects. As shown in Figure~\ref{fig:rebuttal_fig_1}, our method improves upon the upgraded KDAD baseline by producing more semantically meaningful results, with a focus on abnormal objects rather than the entire image.

\end{document}